\documentclass[runningheads]{llncs}

\usepackage{eccv}
\usepackage{pifont} 
\usepackage{newtxmath}
\usepackage{dsfont}
\usepackage{colortbl} 
\usepackage{graphicx}
\usepackage{sidecap}

\usepackage[marginal]{footmisc}

\usepackage{eccvabbrv}

\usepackage{graphicx}
\usepackage{booktabs}

\usepackage[accsupp]{axessibility}  

\usepackage[pagebackref,breaklinks,colorlinks,citecolor=eccvblue]{hyperref}

\usepackage{orcidlink}
\usepackage{multirow}
\usepackage{hyperref}
\usepackage[marginal]{footmisc}
\usepackage[utf8]{inputenc}
\usepackage[T1]{fontenc}

\begin{document}

\title{MLS-Track: Multilevel Semantic Interaction in RMOT} 

\titlerunning{Abbreviated paper title}

\author{Zeliang Ma\inst{1} \and Song Yang\inst{1} \and Zhe Cui\inst{1,2} \and
Zhicheng Zhao\inst{1,2} \and Fei Su\inst{1,2} \and Delong Liu\inst{1} \and Jingyu Wang\inst{3}}
\institute{Beijing University of Posts and Telecommunications \and
Beijing Key Laboratory of Network System and Network Culture, China \and McGill University}
\maketitle


\begin{abstract}
 The new trend in multi-object tracking task is to track objects of interest using natural language. However, the scarcity of paired prompt-instance data hinders its progress. To address this challenge, we propose a high-quality yet low-cost data generation method base on Unreal Engine 5 and construct a brand-new benchmark dataset, named Refer-UE-City, which primarily includes scenes from intersection surveillance videos, detailing the appearance and actions of people and vehicles. Specifically, it provides 14 videos with a total of 714 expressions, and is comparable in scale to the Refer-KITTI dataset. Additionally, we propose a multi-level semantic-guided multi-object framework called \textbf{MLS-Track}, where the interaction between the model and text is enhanced layer by layer through the introduction of Semantic Guidance Module (SGM) and Semantic Correlation Branch (SCB). Extensive experiments on Refer-UE-City and Refer-KITTI datasets demonstrate the effectiveness of our proposed framework and it achieves state-of-the-art performance. Code and datatsets will be available.
\end{abstract}

\section{Introduction}
\label{sec:intro}

In recent years, leveraging natural language descriptions for visual tasks\cite{talk2car,DAVIS,refcoco,Nagaraja,type_to_track,clip} has emerged as a prominent research direction in computer vision. The Referring Multi-Object Tracking (RMOT)\cite{RMOT} task has attracted widespread interest, aiming to track specific objects in images or videos based on human language descriptions. RMOT holds tremendous potential in fields such as intelligent security, autonomous driving, and video editing. Several benchmark datasets\cite{OTB,city-ref,talk2car,DAVIS,refer-youtube} have been released to propel research in this area. Refer-KITTI\cite{RMOT} and GroOT\cite{type_to_track}, focusing on 2D scenes, have greatly facilitated progress in RMOT. The recent introduction of the NuPrompt\cite{prompt} dataset, centered on 3D images, further broadens the potential applications of RMOT in the field of autonomous driving.

These benchmark datasets are typically built upon existing publicly available multi-object tracking datasets\cite{mot17,tao,mot20,3DT,BDD,AVA}. For instance, Refer-KITTI\cite{RMOT} is constructed based on the KITTI\cite{kitti} dataset, where multiple objects meeting text requirements in the videos are labeled through natural language guidance. However, the accuracy of such benchmark datasets is constrained. Firstly, the accuracy of the original dataset has inevitably influenced these benchmark datasets. Secondly, the inconsistency manual annotation standards has also extently affected their accuracy. Moreover, the current textual descriptions in benchmarks primarily focus on the static characteristics of objects, such as color and orientation. They overlook the dynamic features, thus not fully leveraging the characteristics of multi-object tracking tasks.
\begin{figure}[tb]
  \centering
  \includegraphics[width=\linewidth]{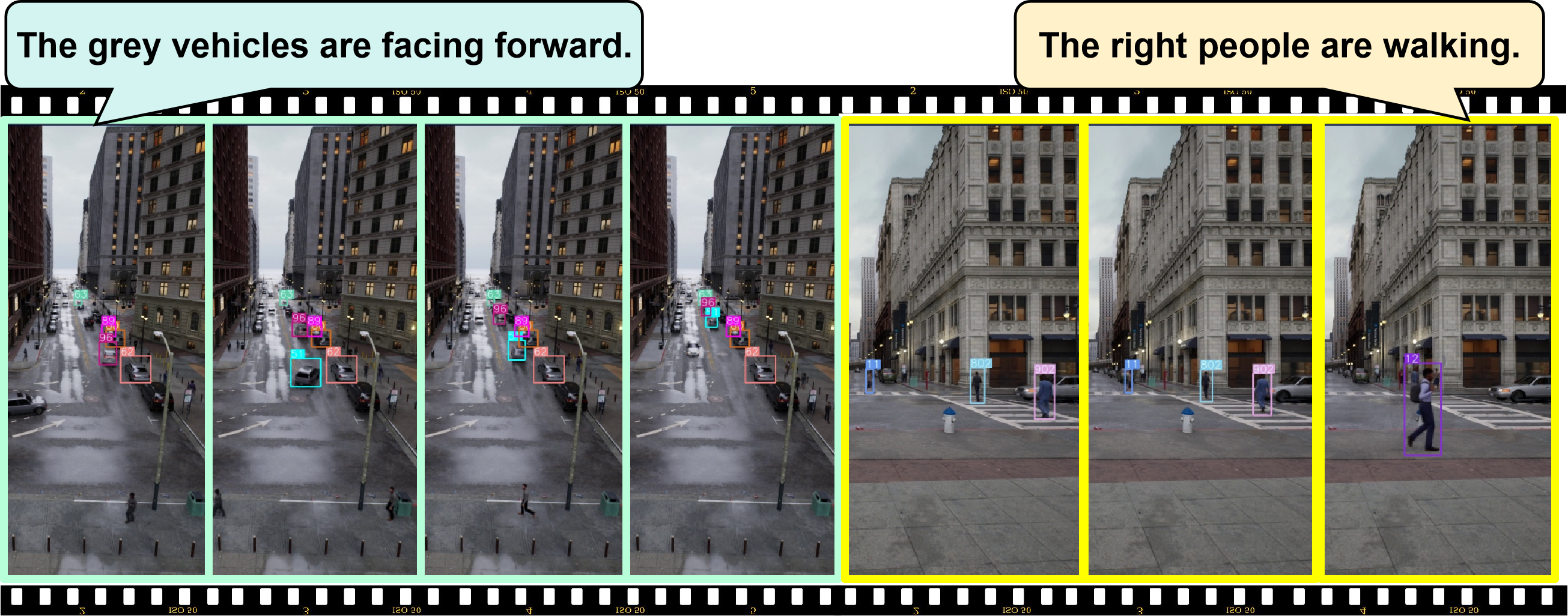}
  \caption{\textbf{Some examples of the proposed Refer-UE-City dataset.} It provides high-quality multi-object annotations for each scene based on different language prompt. Each box color represents a unique identity.
  }
  \label{fig:fig1}
\end{figure}

To address these issues, we propose a new benchmark, named Refer-UE-City. This work utilizes Unreal Engine 5 (UE5) to generate a virtual world and simulates pedestrian and vehicular traffic using internal components. Subsequently, virtual cameras are used to capture video footage while automatically generating 2D bounding box information for pedestrians and vehicles in the scene. Finally, language prompts are assigned to sets of targets. This work simulates the perspective of surveillance cameras, further expanding the application prospects of semantic multi-object tracking in intelligent security. 

Essentially, this benchmark provides instance-text pairings with three primary attributes: \ding{172} \textbf{More dynamic descriptors.} The prompts in our dataset describe various states of vehicle movement at intersections, such as "moving vehicles", "vehicles making turns", and "vehicles facing forward", as shown in \cref{fig:fig1}.  \ding{173} \textbf{More objective descriptors.} Instances are modeled with appearance attributes, eliminating the need for manual annotation of position and color information.  \ding{174} \textbf{Minimal manpower costs.} The stages of dataset creation can be automatically generated, reducing labor costs.

The RMOT task presents additional challenges compared to traditional MOT tasks\cite{mot17,TOP,fairmot,deepsort}, particularly in terms of cross-modal semantic understanding and inter-frame temporal association. To address these challenges, we propose a multi-level semantic-guided framework, named \textbf{MLS-Track}. This framework integrates semantic information layer by layer, from the encoder to prediction head. Specifically, \textbf{MLS-Track} incorporates text features into visual features in the encoder through an early fusion module. In the decoder, it employs a semantic guidance module to facilitate interaction between queries and text. Additionally, drawing inspiration from CLIP\cite{clip}, we add a semantic correlation branch to map features from the fusion space to the text space.

In summary, our contributions are threefold:

1. We introduce a high-quality yet low-cost synthetic dataset generation method based on Unreal Engine 5, along with the construction of a new benchmark dataset called \textbf{Refer-UE-City}. This alleviates the large-scale annotation challenges in refering multi-object tracking task.

2. We propose an end-to-end referring multi-object tracking framework, named \textbf{MLS-Track}. Semantic information from text is effectively integrated into the visual features of queries, enhancing the network's ability to learn from cross-modal data. We devise a novel Semantic Correlation Branch, which enhances the model's generalization ability significantly.

3. Testing on two RMOT datasets demonstrates that the proposed \textbf{MLS-Track} outperforms all state-of-the-art methods, confirming its effectiveness. Furthermore, additional ablation experiments confirm the effectiveness of the network model.

\section{Related Work}

\subsection{Referring Understanding Datasets}
With the continuous development of multimodal tasks, language prompt modeling have gradually encompassed various visual tasks, including object detection\cite{MDETR,MCRE}, segmentation\cite{Cross-Modal}, and target tracking\cite{RMOT,type_to_track}. The introduction of many datasets has also greatly facilitated the advancement of these techniques. Initially, datasets such as Flickr30k\cite{Flickr30k}, ReferIt\cite{ReferIt}, and RefCOCO/g\cite{refcoco} pioneered the task of language-guided object detection by associating linguistic expressions with visual regions in images. However, these datasets were entirely
image-based and cannot adapt well to common video scenarios.

Subsequently, datasets like LaSOT\cite{LaSOT} and TNL2k\cite{TNL2k} introduced textual cues into single-object tracking tasks, annotating not only the position records of targets in each video sequence but also providing language descriptions regarding their appearance. Nevertheless, these datasets were only suitable for expressions of a single target and overlooked temporal variations in targets, such as changes in appearance or behavior over time. 

To address this issue, Refer-KITTI\cite{RMOT} proposed the RMOT task, annotating the set of referred objects for each prompt based on the KITTI\cite{kitti} dataset. However, its accuracy is not only limited by the source dataset but also affected by subjective differences in manual annotation. To alleviate these problems, we adopted UE5 to generate the original tracking dataset and provided finer-grained matching for each example and prompt. \cref{tab:tab1} summarizes a comprehensive comparison between existing prompt-based datasets and our dataset.



\begin{table}[tb]
\fontsize{7}{10}\selectfont
  \caption{\textbf{Comparison of Refer-UE-City with existing datasets.}}
  \label{tab:tab1}
  \centering
  \begin{tabular}{c|c|c|c|c|c|c}
   \toprule[1pt]
    \textbf{Datasets} & \textbf{Basic} \textbf{Task} & \textbf{Frames} & \textbf{Prompt} & \textbf{AnnBoxes}  &\begin{tabular}[c]{@{}c@{}}\textbf{Instances}\\ \textbf{per-prompt}\end{tabular} & \begin{tabular}[c]{@{}c@{}}\textbf{Temporal ratio}\\ \textbf{per-expression}\end{tabular}\\

    \hline
    RefCOCO\cite{refcoco}  & Det\&Seg & 26711 & 142209 & \text{-} & 1&1  \\
    Talk2Car\cite{talk2car} & Det & 9217 & 11959 & \text{-} &1 & \text{-} \\
    Cityscapes-Ref\cite{city-ref} & Det & 4818 & 30000 & \text{-} &1 &1 \\
    LaSOT\cite{LaSOT} & SOT & 3.52M & 10.8k & 3.52M & 1& 1\\
    TNL2K\cite{TNL2k} & SOT & 1.24M & 10.3K & 1.24M &1 &1 \\
    \hline
     Refer-KITTI\cite{RMOT} & RMOT & 6650 &818 &0.35M &10.7 &0.49\\
 \textbf{Refer-UE-City} & \textbf{RMOT} & \textbf{6207} & \textbf{714} & \textbf{0.55M} & \textbf{10.3} & \textbf{0.78}\\
    \bottomrule[1pt]
  \end{tabular}
\end{table}

\subsection{Referring Understanding Methods}
The core challenge of referring understanding is how to model the semantic alignment of cross-modal sources. Benefiting from the flexibility of Transformer\cite{trackformer}, excellent recent solutions\cite{RMOT,type_to_track} mainly follow the joint tracking paradigm. They have made improvements in cross-modal fusion methods to better capture candidate targets in images.

VLTVG\cite{VLTVG} enhanced object recognition performance by concentrating encoded features on language-relevant regions through its Visual-Linguistic Verification Module, while aggregating crucial visual context using the Language-guided Context Encoder. In contrast, the JointNLT\cite{joint} model for single-object semantic tracking employed a multi-relational modeling module to capture relationships between images and different texts, enhancing adaptability to target variations through a semantic-guided temporal modeling module. Nevertheless, the multi-relational modeling module simply stacked self-attention\cite{vit} structures, resulting in both computational resource expenditure and inadequate fusion effectiveness. 

TransRMOT\cite{RMOT} for multi-object semantic tracking proposed an early fusion module, integrating preliminary encoded images and texts through a series of deformable encoder layers. Although this method significantly reduced computational costs, it only performed fusion at the low-level features, leading to insufficient fusion effectiveness. IKUN\cite{ikun} proposed a knowledge unification module (KUM) to adaptively extract visual features based on textual guidance. Its performance is constrained by the tracker, as it treats the RMOT task as a matching problem between trajectories and text.

In comparison to these approaches, we propose a simple cross-modal decoder that combined visual and textual information at higher-level features, resulting in not only lower computational costs but also more effective fusion.

\section{Dataset Overview}

\subsection{Data Collection and Annotation}
\label{sec:section3.1}
\begin{figure}[tb]
  \centering
  \includegraphics[width=\linewidth]{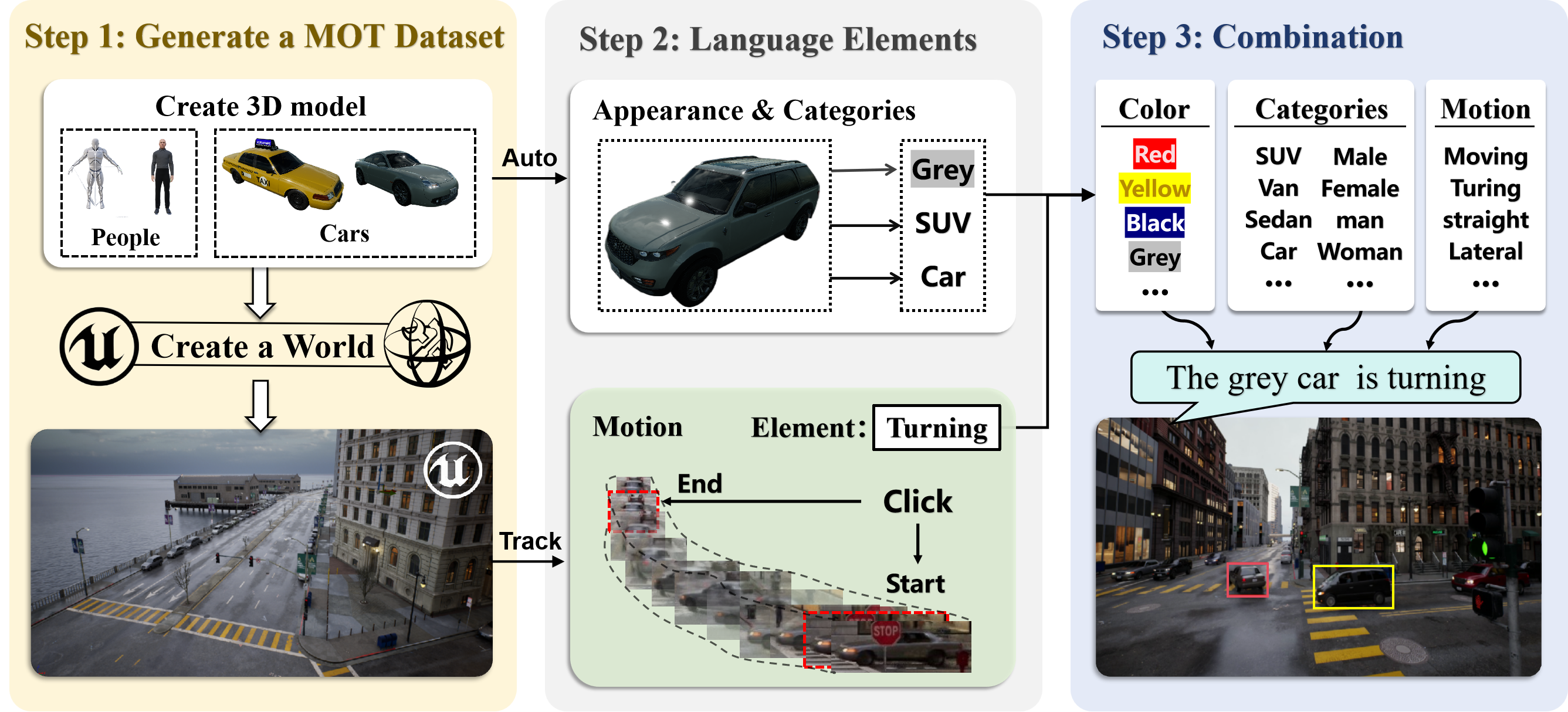}
  \caption{\textbf{Pipeline for generating a language prompt dataset}, consisting of three steps: generate a MOT dataset, identifying language elements, and combination. Firstly, we utilize UE5 to generate a virtual world and record labels for each trajectory within it. Secondly, 3D models automatically output appearance and category information, while we manually annotate their motion information. Lastly, textual descriptions are created through the combination of Language Elements.
  }
  \label{fig:fig3.1}
\end{figure}

Currently, datasets for Referring Multi-Object Tracking are primarily built upon publicly available MOT datasets. These datasets provide a unique identifier for each instance, facilitating annotators to continue constructing RMOT datasets. However, their accuracy is simultaneously constrained by both the accuracy of the original dataset and the inconsistency in manual annotations. Relying solely on 2D visual images may not accurately annotate certain information, posing challenges that interfere with long-term research. In order to effectively construct accurate RMOT datasets, we designed a three-step semi-automatic generation pipeline, as illustrated in \cref{fig:fig3.1}. 

\textbf{Step 1}: Generate a virtual multi-object tracking dataset. Initially, we utilize publicly available resources in UE5 and modify the appearance of 3D models to generate instances of vehicles and pedestrians in the system. Then, these instances are bound with motion events in the traffic system, such as linking vehicles with intersection turning events. Finally, cameras are positioned in the world to record video footage, from which information regarding the bounding box of the targets in the scene can be directly obtained. We have generated the UE-City dataset at a lower cost by continuously altering the 3D modeling of targets and the position of the camera. 

\textbf{Step 2}: Identifying language elements. Since the targets are generated based on predefined configurations, their appearance descriptions and categorized attributes can be automatically obtained without the need for manual annotation. In this step, the manual annotation process solely focuses on describing motion attributes. \cref{fig:fig3.1} illustrates an annotated instance of a car undergoing a turning action. To specify the initiation of the turning, one can simply click on the car's boundary. Similarly, to indicate the conclusion of the turning action, clicking on the car's boundary again is sufficient. The annotation tool automatically populates the labels between the starting and ending frames. Ultimately, save the label information (i.e., frame ID, object ID, and box coordinates), and preserve the corresponding expressions.

\textbf{Step 3}: Language Elements Combination. We employ two relationships to combine language attributes of objects: AND and OR. In this step, we manually select meaningful combinations of attributes and generate a large number of combinations for objects randomly. To ensure the validity of these combinations, we exclude those with quantities below a certain threshold. 

Ultimately, through this process, semantic annotations for the Refer-UE-City dataset are generated. This approach significantly reduces the cost of manual annotation while ensuring the accuracy and objectivity of the dataset.
\begin{figure}[tb]
  \centering
  \includegraphics[width=\linewidth]{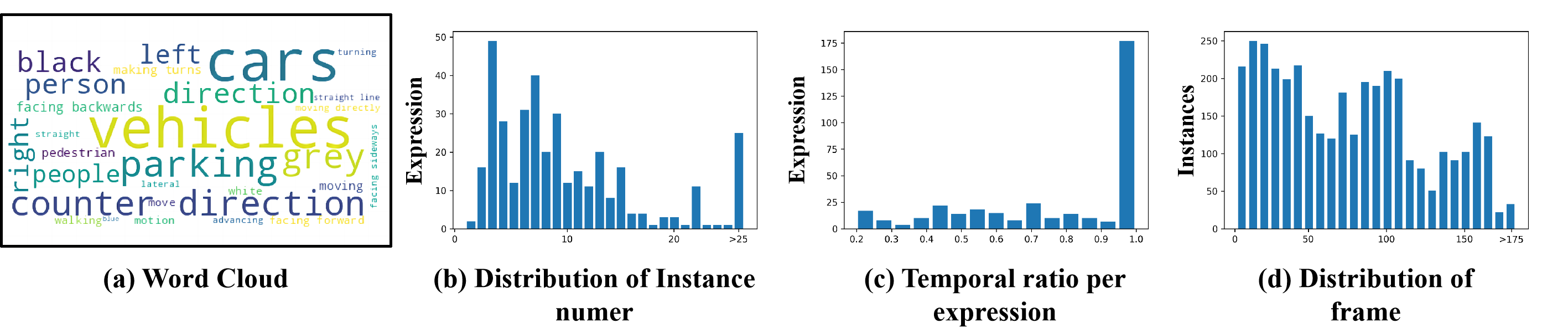}
    \caption{\textbf{Statistics of Refer-UE-City} on (a) word cloud and (b) distribution of Instance number per expression.(c) Distribution of the ratio of referent frames covering video.(d) Distribution of the duration of each referenced instances.}
  \label{fig:fig3.2}
\end{figure}
\subsection{Dataset Statistics}
\label{sec:section3.2}
Refer-UE-City includes various scenes such as urban intersections, traffic roads, and highways. The language prompts offer a wealth of comprehensive descriptions, covering a wide range of objects in both temporal and spatial dimensions. To gain a deeper understanding of Refer-UE-City, we will next discuss the statistical data regarding its language prompts and referent objects.

\textbf{Language prompts}. We manually annotated 716 language prompts, with an average of 51 prompts per video. As shown in \cref{tab:tab1}, Refer-UE-City is comparable to Refer-KITTI in terms of prompt counts. Unlike Refer-KITTI, our dataset is captured from static cameras and our language prompts are more focused on describing vehicle situations at intersections. As depicted in the word cloud in \cref{fig:fig3.2}(a), the Refer-UE-City dataset contains a large number of words describing vehicle actions such as "moving", "turning", "parking" as well as many words describing vehicle appearances like "black", "grey". Additionally, the dataset includes descriptions of vehicle postures, such as "counter direction", "lateral". 

\textbf{Referent objects}. Unlike previous datasets where each language expression pertained to a single specified object, Refer-UE-City aims to cover predictions for any number of objects in the video. As shown in \cref{fig:fig3.2}(b), expressions in Refer-UE-City primarily describe between 1 to 25 instances, with the maximum count reaching up to 54 instances. On average, each expression in the video contains approximately 10.3 instances. In \cref{fig:fig3.2}(c), we demonstrate the proportion of expressions covering the entire video duration. The results show that the majority of language prompts cover the entire duration of the video. \cref{fig:fig3.2}(d) displays the distribution of the duration of each referenced object. The durations of referenced objects vary, with the maximum number of frames exceeding 175 frames, presenting additional challenges for long-term tracking.

\subsection{Benchmark Protocols}
\label{sec:section3.3}
To assess the similarity between predicted trajectories and ground truth, we primarily employ the High-Order Tracking Accuracy (HOTA)\cite{HOTA} as the standard metric. The computation of HOTA involves multiple components, and a simplified calculation formula is provided below:
\begin{align}HOTA_{\alpha}=\sqrt{\frac{\sum_{c\in\{TP\}}\mathcal{A}\left(c\right)}{TP+FN+FP}}=\sqrt{\mathrm{DetA}_{\alpha}\cdot\mathrm{Ass}_{\alpha}}\end{align}
where $\mathcal{A}$ represents the data association score, which measures the effectiveness of target tracking. The concepts of True Positive(TP), False Negative(FN), and False Positive(FP) are the same as those in the MOTA \cite{MOTA}, used to measure the effectiveness of target detection. For the RMOT task, when predicting objects that are visible but not referred to, they are considered false positives, which affects the HOTA metric. In this work, the overall HOTA is calculated by averaging the results of different language prompts.

\section{Method}
\label{sec:blind}
\begin{figure}[tb]
  \centering
  \includegraphics[height=5.1cm]{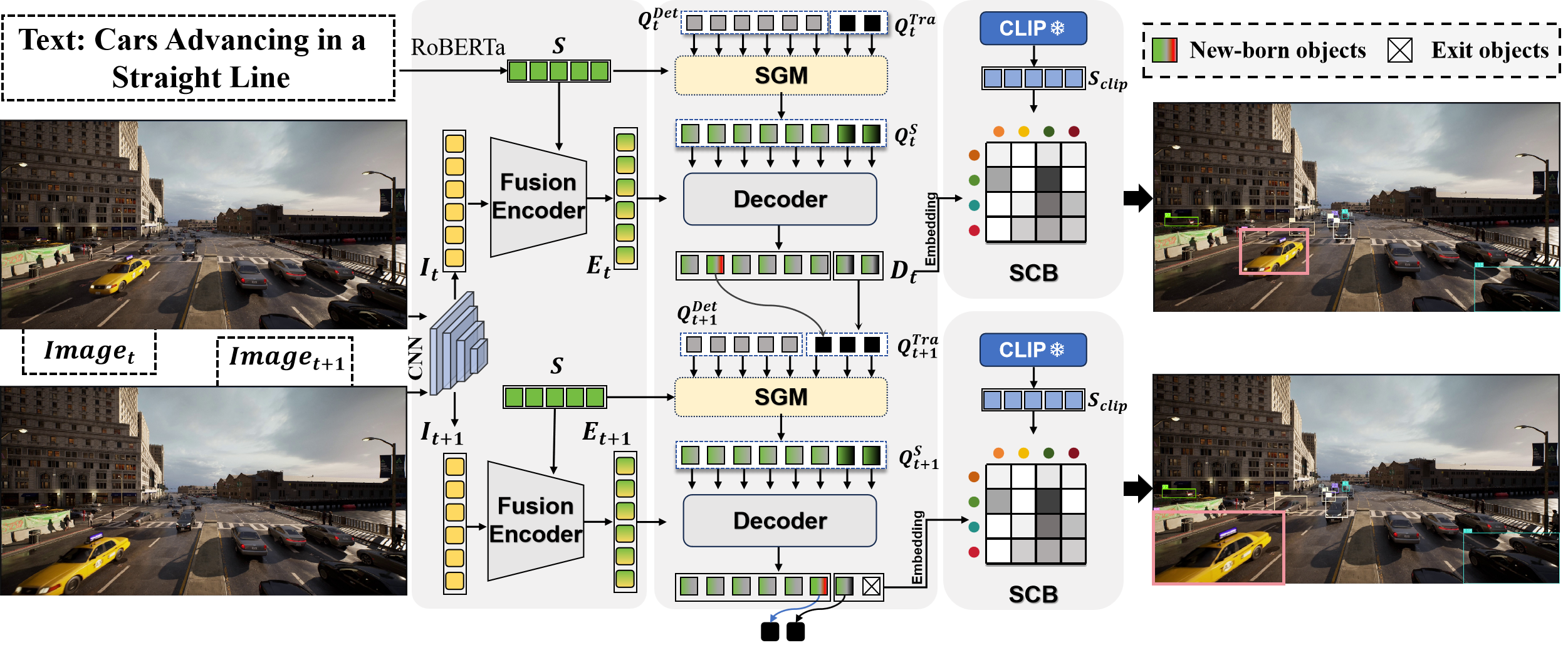}
  \caption{ \textbf{The overall architecture of MLS-Track.} It is an online cross-modal tracker that interacts with semantics layer by layer. Firstly, Fusion Encode conducts early fusion of visual and textual features. Before entering the decoder layer, the Semantic Guidance module embeds language features into query. Subsequently, semantic queries decode cross-modal embeddings. Finally, the queries are projected into the textual space to measure their similarity with the encoded textual features of clips, facilitating the prediction of referenced objects.
  }
  \label{fig:fig4.1}
\end{figure}

Given a language expression as a reference, RMOT targets to ground all semantically matched objects in a video. This task requires not only addressing inter-frame associations but also achieving comprehensive alignment between textual and visual features. To accomplish these objectives, we propose \textbf{MLS-Track}, an end-to-end framework. Building upon the TransRMOT\cite{RMOT} , MLS-Track enhances it by introducing a Semantic Guidance Module and a Semantic Correlation Branch. The overall pipeline of our method is illustrated in \cref{fig:fig4.1}. In the following sections, we will provide a detailed overview of MLS-Track.

\subsection{Overall Architecture}
Our model mainly follows the Deformable DETR\cite{deformable}. We adapt it to the RMOT task through multi-level semantic interactions. Following TransRMOT, we process a T-frame video by utilizing a CNN backbone to extract pyramid feature maps, e.g., the $t^{th}$ frame is represented by features $I_t$. Concurrently, a pre-trained language model is employed to embed text words into vectors $S$. Each layer of the pyramid feature maps $I_t$ interacts with language features through a cross-modal encoder, yielding cross-modal embeddings $E_t$.

Before entering the decoder layers, a semantic guidance module is designed to interact the embeddings of queries with linguistic features, outputting queries embedded with linguistic embeddings as $Q_{t}^s$. Then, $Q_{t}^s$ through a series of Transformer decoder layers interact with cross-modal embeddings $E_t$, embeddings for the decoder are generated. To associate objects between adjacent frames, track queries from the current frame are updated to those of the next frame to continuously track the same instances (${Q_t^{Tra}}\to{Q_{t+1}^{Tra}}$). During this process, linguistic features provide guidance to the embedding of queries, aiding in filtering out semantically irrelevant parts. 

After a group of decoder layers, in addition to the category and bounding box prediction branches, we also add a semantic correlation branch on top of the decoder. On this branch, leveraging the powerful text-image alignment capability of CLIP, the text is re-encoded to obtain $S_{clip}$. Then, the similarity between $S_{clip}$ and the query embedding is computed. Finally, predictions with low similarity are filtered out to obtain reference objects that align with the semantics of the current frame.

The entire tracker is optimized with a multi-frame loss function. The loss function for each frame is formulated as: $\mathcal{L} = \lambda_{cls}\mathcal{L}_{cls}+\lambda_{L_1}\mathcal{L}_{L_1}+\lambda_{g_{iou}}\mathcal{L}_{g_{iou}}+\lambda_{ref}\mathcal{L}_{ref}$, where $\mathcal{L}_{cls}$, $\mathcal{L}_{L_1}$ and $\mathcal{L}_{g_{iou}}$ are the focal loss\cite{focal_loss}, L1 loss and IoU loss. $\lambda_{L_1}$, $\lambda_{cls}$, $\lambda_{g_{iou}}$, $\lambda_{ref}$ are the corresponding hyper-parameters.
\subsection{Semantic Guidance Module}
\begin{figure}[tb]
   \centering
  \includegraphics[width=\linewidth]{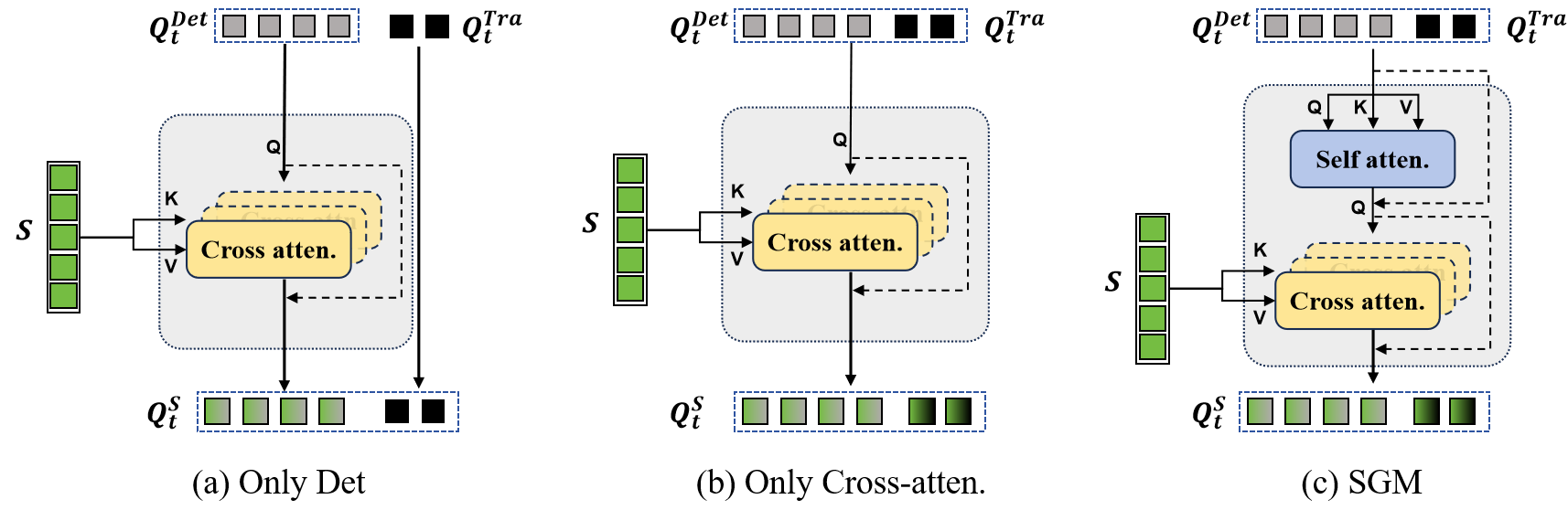}
        \caption{\textbf{Three designs of Semantic Guidance Module.} The dashed line represents the residual structure.}

  \label{fig:fig4.2}
\end{figure}

The early fusion module of TransRMOT\cite{RMOT} focuses on integrating visual and textual features. While this approach effectively integrates textual information into visual features, it also potentially increases the difficulty of decoding fused features. We believe that it is necessary for the query, serving as the carrier of the target in the trajectory, to perceive semantic information in advance. Therefore, we propose a Semantic Guidance Module to guide the query to perceive the semantics of the target before entering the decoder, as illustrated in \cref{fig:fig4.2}(c).

Specifically, at the $t^{th}$ frame, the query includes not only the initialized detect query $Q_t^{Det}$ but also the track query $Q_t^{Tra}$ from the previous frame. In practice, the two kinds of queries are concatenated together, denoted as $Q_t$.

Firstly, we add position embedding to each query feature and use it as the input to the self-attention. The result is then normalized, passed through a feed-forward neural network, and applied residual connections throughout the process, yielding $\widehat{Q}_t$.

Next, the text features are linearly mapped to the same dimensionality as $\widehat{Q}_t$, and then cross-attention is employed to encourage feature fusion between the two modalities $\widehat{Q}_t$ and $S$, resulting in the prompt-aware query $Q_t^s\in\mathbb{R}^{N\times C}$:

\begin{gather}
S_{emb}^{\prime}=Linear(S_{emb})\\
Q_t^s=~CrossAttn(~Q=\widehat{Q}_t~,K=S_{emb}^{\prime},V~=~S_{emb}^{\prime})
\end{gather}
where $\widehat{Q}_t\in\mathbb{R}^{N\times D}$, N represents the number of queries, D is the dimension of features.And $S_{emb}$ indicates semantic features, $S_{emb}\in\mathbb{R}^{L\times D}$, L is the length of words in semantics. Overall, the query undergoes self-attention to ensure specificity between queries and facilitate interaction between $Q^{Det}$ and $Q^{Tra}$, preventing them from competing for the same target. Subsequently, fusion with text features through cross-attention enables early perception of semantic reference targets.

\subsection{Semantic Correlation Branch}
The semantic correlation branch aims to generate binary-valued scores indicating the similarity between trajectories and text. By filtering out targets with low similarity, it can suppress the appearance of objects irrelevant to the text. Unlike the classification prediction branch, a simple FFN struggles to represent the mapping relationship between visual features and semantic features. Therefore, we designed an branch similar to CLIP\cite{clip} to achieve semantic relevance prediction through cross-modal alignment.

Specifically, we map the query embedding to the semantic feature space through a linear mapping and calculate its cosine similarity with the textual features to obtain reference scores:

\begin{gather}
\hat{S}_{emb}=Encoder_{clip}(\mathcal{S}_{emb})\\
\hat{D}_t=Linear(D_t)\\
R_{cos}=Cos\_distance(\hat{S}_{emb},\hat{D}_t)
\end{gather}
where $D_t$ represents the output of $Q_{t}^{s}$ after passing through the decoder, and $Encoder_{clip}$ denotes the text encoder of CLIP\cite{clip}.

\section{Experiments}

\subsection{Implementation Details}
\textbf{Dataset.} The training set of Refer-UE-City comprises 10 videos with 96 distinct descriptions, while the test set consists of 2 videos with 52 unique descriptions. To evaluate the method's generalization, we introduce Refer-KITTI, which has 15 videos in the training set and 3 videos in the test set. All ablation experiments are conducted on these two datasets to ensure the model's generalization.\\
\textbf{Model Details.} Following TransRMOT, we use ResNet-50\cite{resnet} to extract visual features and RoBERTa\cite{roberta} to embed language prompts. In the semantic guidance module, the number of attention heads for multi-head attention is set to 8, and the output feature dimension is set to 256. The number of queries is set to N = 300. In the prompt probability prediction branch, the pre-trained model for CLIP is set to ViT-Large model.\\
\textbf{Training and Testing.} During the training phase, both RoBERTa\cite{roberta} and CLIP\cite{clip} are frozen. The initial learning rate for the model backbone is set to 1e-5, while other parts are set to 1e-4. The model is trained for 60 epochs. It's worth noting that during training, to fully unleash the performance of the Transformer, we increase the Track Query to 3 groups following MOTRv3\cite{motrv3}. Additionally, Gaussian noise is added to the embeddings with a mean and variance of 0 and 0.3. However, during the inference phase, we discard the Track Group Denoising. The overall training is deployed on 2 RTX 4090 GPUs with batch size of 1.

\subsection{Benchmark Experiments}
We compare the experimental results of our method with existing methods on Refer-KITTI and Refer-UE-City datasets as shown in \cref{tab:tab5.2}. The performance of the current state-of-the-art end-to-end method, TransRMOT, on the Refer-KITTI dataset is 38.06\% (HOTA). During our reimplementation of this method, we uncovered a bug in the code where an offset existed between predicted frames and actual frames. By fixing this issue, the results of the corrected version (TransRMOT$\ast$) showed significant improvement. \textbf{MLS-Track} outperforms it by 9.84\% HOTA on the Refer-UE-City dataset and by 3.4\% HOTA on the Refer-KITTI dataset.

Moreover, we construct a series of CNN-based competitors by integrating our cross-modal fusion module into the detection part of multi-object tracking models. IKUN\cite{ikun} is a two-stage framework that first tracks all targets and then matches trajectories with semantics. In both datasets, \textbf{MLS-Track} outperforms these two-stage networks, further demonstrating the effectiveness and generality of our approach. This also underscores that relying solely on single-frame grounding is insufficient for matching action semantics.

\begin{table}[tb]
\fontsize{8}{12}\selectfont
\caption{\textbf{Comparison with state-of-the-art RMOT methods} on Refer-KITTI(R-KITTI) and Quantitative results on Refer-UE-City(R-city). "E" means "End to end" , TransRMOT$\ast$ means the TransRMOT result after frame correction. $\ddag$: the detection results are from cross-model Deformable DETR\cite{deformable}.
}
\label{tab:tab5.2}
\begin{center}
\setlength{\tabcolsep}{0.8mm}{
\begin{tabular}{c|c|c|c|c|c|c|c|c|c}
\toprule[1pt]
\textbf{Method} & \textbf{Datasets} & \textbf{E} & \textbf{HOTA} & \textbf{DetA} & \textbf{AssA} & \textbf{DetRe} & \textbf{DetPr} & \textbf{AssRe} &\textbf{AssPr}\\
\hline
DeepSORT$\ddag$\cite{deepsort} & R-city & $\times$&24.34&14.35&42.46&19.99&31.09&49.23&69.95 \\
ByteTrack$\ddag$\cite{bytetrack} & R-city & $\times$ &23.22&13.99&39.62&19.25&31.19&45.91&69.72 \\
StrongSORT$\ddag$\cite{strongsort} & R-city &$\times$ & 24.43 &14.38& 42.72 & 20.05& 31.09& 50.13& 69.01 \\
iKUN$\ddag$\cite{ikun} & R-city & $\times$ & 10.97 & 4.64 & 26.51 & 5.86 &16.83 & 33.15 &47.52\\
\hline
TransRMOT$\ast$ & R-city & \textbf{\checkmark} & 18.61 & 8.95 & 40.15 & 16.40 & 15.75 & 49.19 & 65.22\\
\textbf{MLS-Track(ours)} & R-city & \textbf{\checkmark} & \textbf{28.45} & \textbf{15.88} & \textbf{53.17} & \textbf{26.14} & \textbf{26.87} & \textbf{60.59} & \textbf{73.99}\\
\hline
FairMOT\cite{fairmot} & R-KITTI & $\times$ & 23.46 & 14.84 & 40.15 & 17.40 & 43.58 & 53.35 & 73.15\\
DeepSORT\cite{deepsort} & R-KITTI & $\times$ & 25.59 & 19.76 & 34.31 & 26.38 & 36.93 & 39.55 & 61.05\\
ByteTrack\cite{bytetrack} & R-KITTI & $\times$ & 22.49 & 13.17 & 40.62 & 16.13 & 36.61 & 46.09 & 73.39\\
CSTrack\cite{CSTrack} & R-KITTI & $\times$ & 27.91 & 20.65 & 39.10 & 33.76 & 32.61 & 43.12 & 71.82\\
TransTrack\cite{transtrack} & R-KITTI & $\times$ & 32.77 & 23.31 & 45.71 & 32.33 & 42.23 & 49.99 & 78.74\\
TrackFormer\cite{trackformer} & R-KITTI & $\times$ & 33.26 & 25.44 & 45.87 & 35.21 & 42.19 & 50.26 & 78.92\\
iKUN\cite{ikun} & R-KITTI & $\times$ & 48.84 & 35.74 & 66.80 & 51.97 & 52.26 & 72.95& 87.09\\
\hline
TransRMOT\cite{RMOT} & R-KITTI & \textbf{\checkmark} & 38.06 & 29.28 & 50.83 & 40.20 & 47.36 & 55.43 & 81.36\\
TransRMOT$\ast$ & R-KITTI & \textbf{\checkmark} & 45.65 & 36.15 & 57.86 & 54.58 & 50.65 & 61.15 & 89.96\\
\textbf{MLS-Track(ours)} & R-KITTI & \checkmark & \textbf{49.05} & \textbf{40.03} & \textbf{60.25} & \textbf{59.07} & \textbf{54.18} & \textbf{65.12} & \textbf{88.12}\\
\bottomrule[1pt]
\end{tabular}}
\end{center}
\end{table}

\begin{table}[tb]
\fontsize{8}{12}\selectfont
  \caption{\textbf{Ablation studies of different components} in Refer-KITTI and Refer-UE-City. Parameters represents the number of learnable parameters in the corresponding model.
  }
  \label{tab:tab5.3.0}
\begin{center}
\begin{tabular}{c|c|c|c|c|c|c|c}
  \toprule[1pt]
  \textbf{Method} & \textbf{Datasets} & \textbf{SGM} & \textbf{SCB} & \textbf{Parameters} & \textbf{HOTA} & \textbf{DetA} & \textbf{AssA} \\
  \hline
  TransRMOT$\ast$ & R-city &  &  & 42.8M & 18.614 & 8.9543 & 40.145\\
    & R-city & \checkmark &  & 43.0M &28.128&15.551&53.372 \\
  \textbf{MLS-Track(ours)} & R-city & \checkmark & \checkmark & \textbf{43.5M} &28.453   &15.88  &53.166  \\
  \hline
  TransRMOT$\ast$ & R-KITTI &  &  & 42.8M & 45.654 & 36.153 & 57.864\\
    & R-KITTI & \checkmark &  & 43.0M & 48.235&	40.071&	58.254\\
  \textbf{MLS-Track(ours)} & R-KITTI & \checkmark & \checkmark & \textbf{43.5M} & \textbf{49.048} & \textbf{40.034} & \textbf{60.254}\\
  \bottomrule[1pt]
\end{tabular}
 \end{center}
 \end{table}
 
\begin{table}[tb]
\fontsize{8}{12}\selectfont
  \caption{\textbf{Ablation study on different components of SGM.} "Only Det" means using Only detect query and text interaction. "Only Cross-atten." means interacting only using cross-attn for all queries and text.}
  \label{tab:tab5.3.1}
\begin{center}
\begin{tabular}{c|c|c|c|c|c|c|c}
\toprule[1pt]
   \multicolumn{1}{c|}{\multirow{2}{*}{\textbf{Method}}} &\multicolumn{1}{c|}{\multirow{2}{*}{\textbf{SGM}}} & \multicolumn{3}{c|}{\textbf{R-KITTI}} & \multicolumn{3}{c}{\textbf{R-city}} \\
   \cline{3-8} 
   \multicolumn{1}{c|}{} & \multicolumn{1}{c|}{} & \textbf{HOTA} & \textbf{DetA} & \textbf{AssA} & \textbf{HOTA} & \textbf{DetA} & \textbf{AssA}\\
   \hline
   TransRMOT$\ast$ & -& 45.654 & 36.153 & 57.864& 18.614 & 8.9543 & 40.145 \\
    & Only Det & 46.874 & 38.386 &57.427 & 23.868 & 12.623 &47.013\\
    & Only Cross-atten. & 48.006&39.907 &57.902& 27.168&14.787 & 51.944\\
   \textbf{Ours} & \textbf{SGM} & \textbf{48.235}&\textbf{40.071} & \textbf{58.254} &\textbf{28.128}&\textbf{15.551}&\textbf{53.372} \\
\bottomrule[1pt]
\end{tabular}
 \end{center}
 \end{table}
\subsection{Ablation Study}
\begin{figure}[tb]
  \centering
  \includegraphics[height=3.8cm]{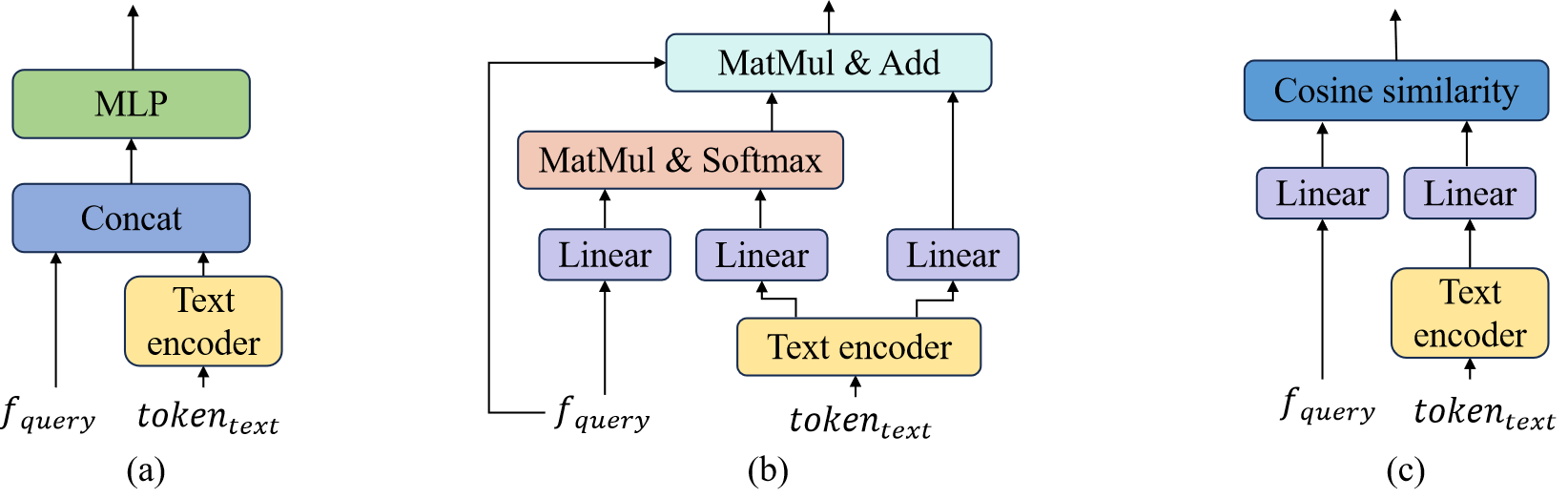}
  \caption{\textbf{Three designs of SCB.} (a) means Concat \& MLP. (b) means cross-attention. (c) is our method.}
  \label{fig:fig5.2}
\end{figure}

\textbf{Semantic Components.} To investigate the impact of core components in our model, we conducted extensive ablation studies on Refer-UE-City and Rerfer-KITTI datasets. According to the results in \cref{tab:tab5.3.0}, the performance on both Refer-KITTI and Refer-UE-City datasets improved by \textbf{2.6\%} and \textbf{9.5\%}, respectively, with the addition of the semantic guidance module. By adding the semantic correlation branch, the performance on Refer-KITTI and Refer-UE-City datasets also improved by \textbf{3.4\%} and \textbf{9.8\%}, respectively. Additionally, by observing the data of the number of trainable parameters in the table, it can be inferred that the two modules only increased the model's parameter count slightly but significantly enhanced the model's performance, thus demonstrating the practicality of the proposed approach.\\
\textbf{Semantic Guidance Module.} As mentioned earlier, the semantic guidance module aims to enable queries to perceive the semantic information of the target in advance. This validates the necessity of the module. The simplest idea is to use text information directly as the initial value of the query, a method that has been proven in previous work\cite{RMOT} to significantly decrease all metrics. The reason is that it did not consider the modality conversion between text and vision and reduced the specificity between queries. We also investigated two different variants in terms of structure, as shown in \cref{fig:fig4.2}. The first variant involves guiding only the detection queries semantically, while the second variant removes self-attention. Our method achieved better results than both variants, as shown in \cref{tab:tab5.3.1}, demonstrating the rationality of our structural design.\\
\textbf{Semantic Correlation Branch.} We explored various methods of the hybrid refer predictor. Firstly, we designed two variants, as depicted in \cref{fig:fig5.2}. The first variant concatenated the text encoding and query features together, then regressed the reference scores through MLP layers. The second one employed cross-attention to fuse the features before regressing the reference scores. However, according to the results shown in \cref{tab:subtable-a}, both of these methods led to a decrease in model performance. This is because our model already integrates visual and text features during the encode and decode stages. In such a scenario, simply adding weighted sums of fused features and text features would bias the final features towards text features. Moreover, as the text features in the structure have not passed through the decode layers, they exhibit significant deviation from the fused features, rendering this interaction meaningless. 

Instead, treating the task as a mapping relationship from the fused feature space to the text space may be a better choice. We also studied different language encoders and compared the performance of RoBERTa and CLIP.  \cref{tab:subtable-a} shows that the CLIP encoder is more suitable for our architecture, achieving an HOTA performance of 49.048\%, compared to RoBERTa's 47.074\%.\\
\textbf{Referring Threshold.}
Finally, we investigated the impact of the reference threshold $\beta_{ref}$ . As shown in \cref{tab:subtable-b}, we sampled the range from 0.3 to 0.6, and the fluctuation range of HOTA was within 0.02\%. Overall, the referring performance is robust to the varying reference threshold. We chose $\beta_{ref}=0.5$  as default.

\begin{table}[tb]

  \caption{\textbf{Ablation study on different components} in Refer-KITTI.
  }
  \label{tab:tab5.3.30}
  \begin{subtable}{0.5\linewidth}
\begin{center}
\caption{Comparison on SCB.}
\resizebox{!}{1.00cm}{
\begin{tabular}{c|c|c|c}
  \hline
   \textbf{Method} & \textbf{HOTA} & \textbf{DetA} & \textbf{AssA} \\
   \hline
   Concat \& MLP & 46.552 & 37.934 & 57.345\\
   \hline
   Cross attention &46.4 & 37.809 & 57.105 \\
   \hline
    Contrast-Robert&47.074 & 38.136 & 58.257 \\
    \hline
   Contrast-clip &\textbf{49.048} & \textbf{40.034} & \textbf{60.254}\\
  \hline
\end{tabular}
}
\label{tab:subtable-a}
 \end{center}      
 \end{subtable}
   \begin{subtable}{0.5\linewidth}
\begin{center}
\caption{Comparison on Refer Threshold $\beta_{ref}$}
\resizebox{!}{1.00cm}{
\begin{tabular}{c|c|c|c}
  \hline
   \textbf{Refer Threshold} & \textbf{HOTA} & \textbf{DetA} & \textbf{AssA}\\
   \hline
    0.3&49.032&39.873& \textbf{60.458}\\
    \hline
   0.4 & 49.038 & 39.969 & 60.33 \\
  \hline
  0.5 &\textbf{49.048} & 40.034 & 60.254 \\
  \hline
  0.6 &49.032 & \textbf{40.079} & 60.148 \\
  \hline
\end{tabular}
}
\label{tab:subtable-b}
 \end{center}
 \end{subtable}
 \end{table}
\subsection{Visualization}

\begin{figure}[tb]
  \centering
  \includegraphics[width=\linewidth]{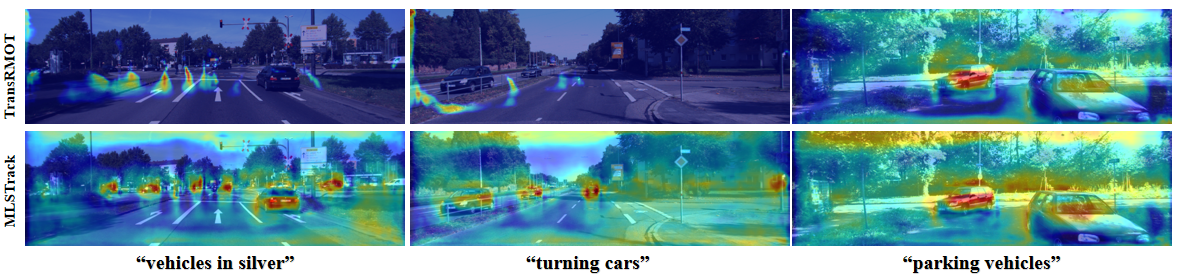}
  \caption{\textbf{Heat map in the last encoder.}
  }
  \label{fig:fig5.4.2}
\end{figure}

\begin{figure}[tb]
  \centering
  \includegraphics[width=\linewidth]{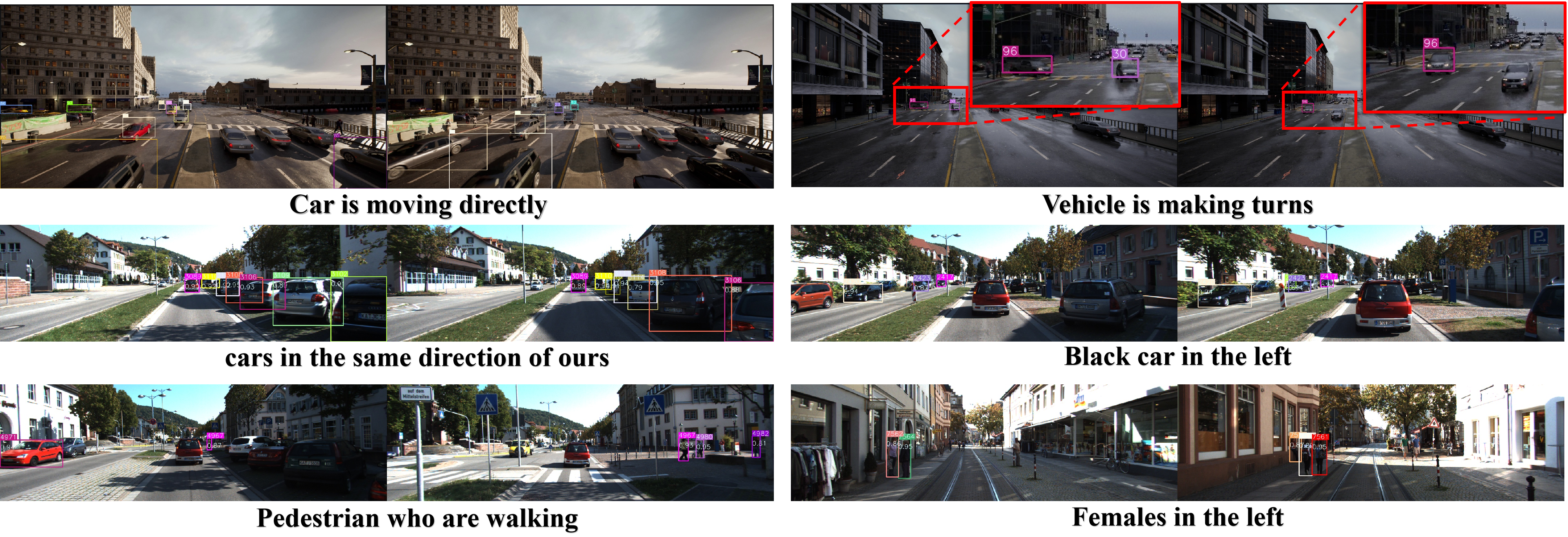}
  \caption{\textbf{Qualitative results of our method on Refer-KITTI and Refer-UE-city.} The first line is from Refer-UE-City, and the rest are from Refer-KITTI.}
  \label{fig:fig5.4.3}
\end{figure}

To illustrate the influence of semantic-guided module backpropagation on the pre-fusion stage, we visualize the heatmap of the last layer of the encoder in \cref{fig:fig5.4.2}. It is shown that the addition of the semantic-guided module tends to make the encoder module focus more on regions close to the reference text. In \cref{fig:fig5.4.3}, we visualize six typical reference results. The results indicate that our method can successfully track targets based on various queries.

\section{Conclusion}
In this work, we presented Refer-UE-City, the first virtual language prompt dataset designed specifically for multi-object tracking in surveillance videos. Compared to previous benchmarks, using virtual engine to generate targets eliminates the need for manual annotation of appearance and categories, resulting in more objective annotations and significantly reducing human labor costs, and allowing for large-scale generation. Additionally, we propose a multi-level semantic-guided  tracking model named \textbf{MLS-Track}. Specifically, we progressively enhance the model's interaction with text by introducing the Semantic Guidance Module (SGM) and Semantic Correlation Branch (SCB). Extensive experiments on the Refer-UE-City and Refer-KITTI datasets validate the effectiveness of our framework, achieving state-of-the-art performance.

\clearpage  

%
%
\bibliographystyle{splncs04}
\bibliography{main}
\end{document}